\documentclass[runningheads]{llncs}

\usepackage{eccv}

\usepackage{eccvabbrv}

\usepackage{graphicx}
\usepackage{booktabs}

\usepackage[accsupp]{axessibility}  

\usepackage{hyperref}

\usepackage{orcidlink}

\usepackage{xcolor}
\usepackage{textcomp}
\usepackage{amsmath}
\usepackage{siunitx} 
\usepackage{algorithm}
\usepackage{algpseudocode}
\usepackage{nicematrix}
\usepackage[dvipsnames]{xcolor} 
\usepackage{colortbl}
\usepackage{multirow}
\usepackage{color}
\usepackage{nicefrac}       
\usepackage{microtype}      
\usepackage{caption}
\usepackage{mathtools}
\usepackage{listings}

\newif\ifdraft
\draftfalse       

\def\paperTitle{Linear Fusion MultiDiffusion for Fast Training-Free Spherical Panorama Generation}
\newcommand{\method}{\textit{LF-MultiDiffusion}\xspace}
\newcommand{\methodfull}{\textit{Linear Fusion MultiDiffusion}\xspace}
\newcommand{\thsi}{360\textdegree\xspace}
\newcommand{\prsp}{\mathcal{I}}
\newcommand{\condprsp}{\mathcal{Y}}
\newcommand{\pano}{\mathcal{J}}
\newcommand{\condpano}{\mathcal{Z}}
\newcommand{\mapset}{\mathcal{S}}
\newcommand{\R}{\mathbb{R}}
\newcommand{\floor}[1]{\lfloor #1 \rfloor}

\newcommand\pmnum[1]{\scriptsize$\pm$#1}
\DeclareMathOperator*{\argmin}{arg\,min}

\newcommand{\beginsupplement}{
    \appendix
	\setcounter{table}{0}
	\renewcommand{\thetable}{A\arabic{table}}%
	\setcounter{figure}{0}
	\renewcommand{\thefigure}{A\arabic{figure}}%
	\setcounter{equation}{0}
	\renewcommand{\theequation}{A\arabic{equation}}
    \setcounter{algorithm}{0}
	\renewcommand{\thealgorithm}{A\arabic{algorithm}}
}

\definecolor{headerColor}{RGB}{224, 224, 224}
\definecolor{highlightColor}{RGB}{230, 244, 252}
\newcommand*{\deemph}[1]{\textcolor{gray}{#1}}

\begin{document}

\title{\paperTitle} 

\titlerunning{LF-MultiDiffusion}

\author{
Akio Hayakawa\inst{1}\orcidlink{0009-0007-5769-5711} 
\and
Yusuke Mukuta\inst{1,2}\orcidlink{0000-0002-7727-5681} 
\and
Tatsuya Harada\inst{1,2}\orcidlink{0000-0002-3712-3691}
}

\authorrunning{
A.~Hayakawa et al.
}

\institute{
The University of Tokyo \and RIKEN Center for Advanced Intelligence Project
\email{\{hayakawa,mukuta,harada\}@mi.t.u-tokyo.ac.jp}
}

\maketitle

\begin{figure}[h]
    \centering
    \includegraphics[width=\linewidth]{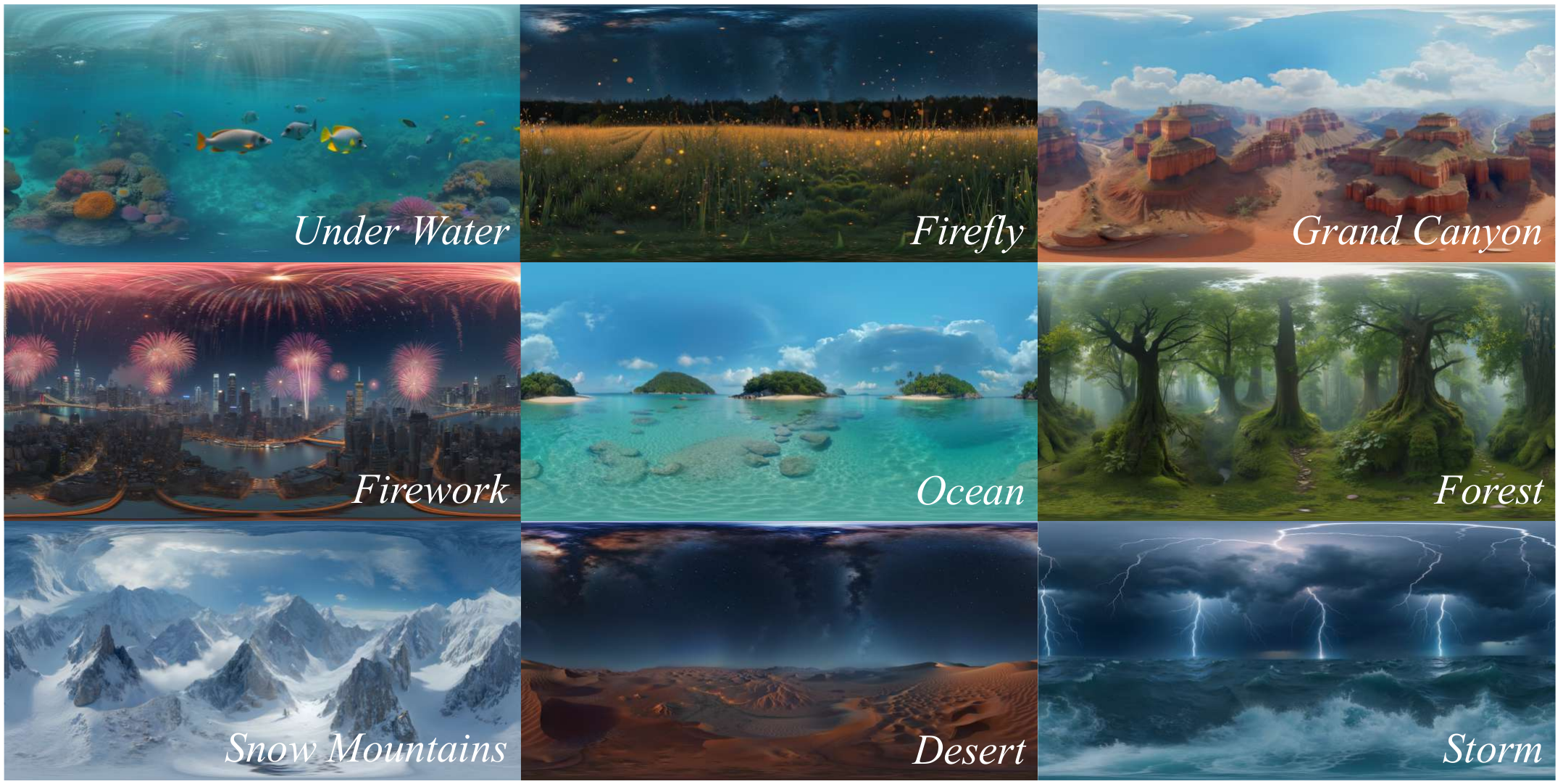}
    \caption{\method enables high-quality panorama generation using a pre-trained text-to-image generator in a training-free manner, achieving SOTA with faster runtime.}
    \label{fig:pano_gallery}
\end{figure}
\begin{abstract}

We propose \method, a training-free panorama generation method that extends MultiDiffusion to support linear projections between target and reference image spaces.
Our key idea is to reformulate latent aggregation as a regularized least-squares problem and solve it efficiently with a Krylov-based iterative solver inside the denoising loop.
This formulation enables denser and more natural mappings than prior training-free methods, yielding more stable generation with far fewer perspective views.
As a result, \method reduces the number of image generator evaluations during denoising and significantly improves inference efficiency.
Experiments show that \method achieves better visual quality, text alignment, and panoramic consistency than the strongest training-free baseline, while providing a 15.36$\times$ speedup.
Our project page is available at: \url{https://ahykw.github.io/lfmd}.


\end{abstract}
\section{Introduction}
\label{sec:intro}
Panoramic imagery has become a fundamental component of immersive digital experiences, including content for head-mounted displays (HMDs), real estate virtual tours, and street-level mapping.
It represents a \thsi scene in a single compact image, enabling users to look around seamlessly.
As commercial HMDs have become widely available, the demand for panoramic content is growing not only for capturing and sharing real-world scenes but also for creating them.  

Recent advances in generative modeling enable high-quality image synthesis \cite{Rombach2022LDM,esser2024scaling, flux2024}, but most models are designed for a limited field of view.
Several works~\cite{tang2023mvdiffusion,Ye2024DiffPano,panfusion2024,kalischek2025cubediff, ni2025UniPano} have extended image generation to panoramic image synthesis using text-ERP paired datasets, but their applicability remains limited to specific domains due to the scarcity of such datasets.
Moreover, there is a substantial gap in domain coverage between general image generation models and panorama-specific models.
For example, image generation models can be trained on diverse visual domains such as artistic or cartoon-style images, whereas such content is rarely available in panoramic datasets.
These limitations make training-free approaches, which fully leverage the capability of pre-trained image generation models, particularly attractive for panoramic content creation.

A few prior works have explored this direction. 
DynamicScaler~\cite{Liu2025DynamicScaler} proposes Panoramic Projecting Denoise, which generates panoramic images by optimizing panorama-domain noisy latents so that perspective patches obtained via equirectangular projection (ERP) follow the reverse diffusion trajectories of a pre-trained image generator.
SphereDiff~\cite{park2025spherediff} instead parameterizes latents directly on the sphere surface to reduce distortion near the pole.
While these methods achieve seamless and high-quality panorama generation, their optimization relies on MultiDiffusion~\cite{Bar-Tal2023MultiDiffusion}, which restricts the projection between the panorama and perspective spaces to direct pixel sampling (or bijective mapping) for each camera view.
As a result, they require a large number of overlapping views (44 in DynamicScaler and 89 in SphereDiff) to cover all pixels in the panorama and maintain seamlessness.
Because a pre-trained image generator must be evaluated for every view at every denoising step, this incurs a high computational cost; generating a single panorama takes about 7 minutes with DynamicScaler and 32 minutes with SphereDiff using \textsc{FLUX} \cite{flux2024}.

\begin{figure}[t]
  \centering
  \includegraphics[width=\linewidth]{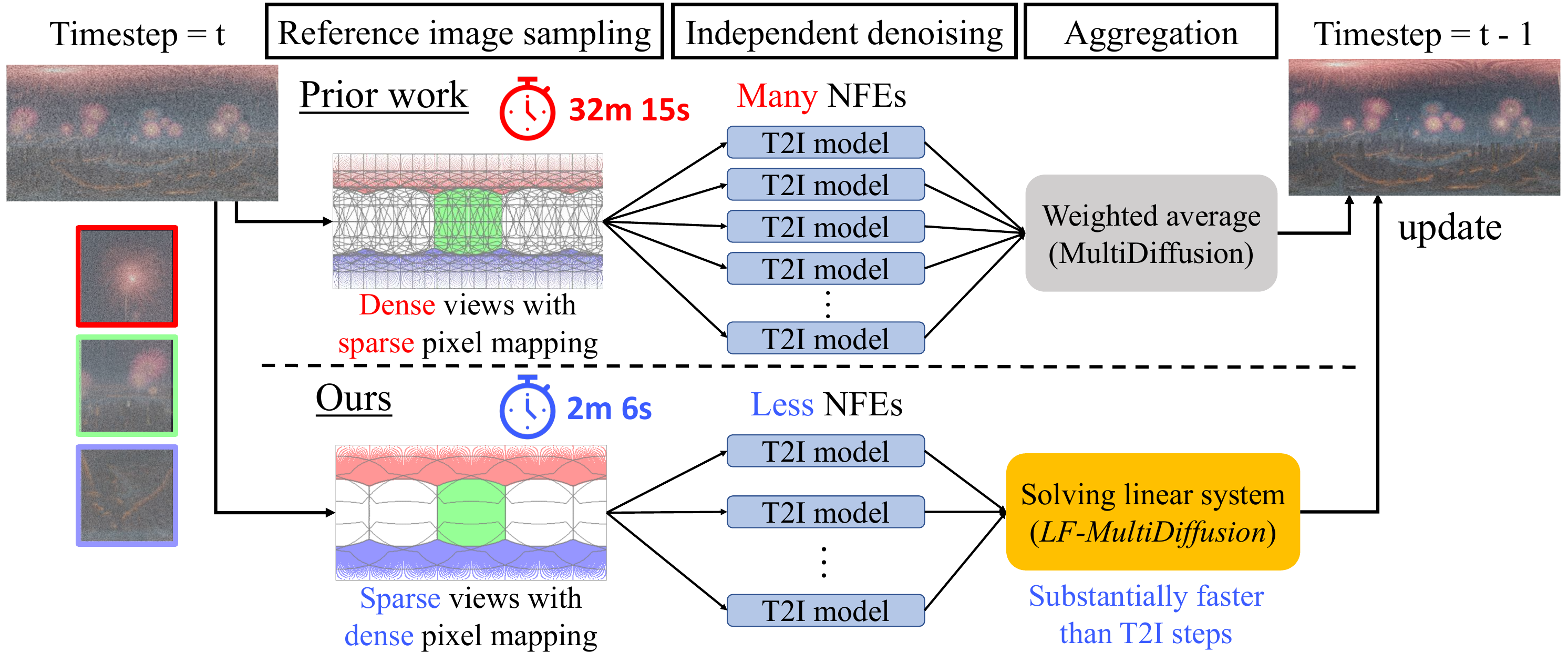}
  \caption{\method for fast training-free panorama generation. Prior methods are constrained to perform direct pixel sampling within the MultiDiffusion framework~\cite{Bar-Tal2023MultiDiffusion}, leading to many neural function evaluations (NFEs) and slow inference. Our method supports denser linear mappings, improving both generation quality and efficiency. Camera frustums and inference times for prior work are taken from SphereDiff~\cite{park2025spherediff}.}
  \label{fig:teaser}
\end{figure}

To address this limitation, we propose \methodfull (\method), an extension of MultiDiffusion that supports arbitrary linear projections between the target and reference image spaces (\cref{fig:teaser}).
Our formulation casts latent aggregation as a regularized least-squares problem.
Its solution is no longer a simple weighted average of reprojected denoised latents, and explicit inversion is computationally impractical.
We therefore incorporate a Krylov-based iterative solver into the denoising loop.
We empirically show that this iterative update is efficient, adding only 20\% of the cost of the denoising step.
By supporting arbitrary linear projections, our method enables denser and more natural mappings between panoramic and perspective views, which stabilizes generation even with far fewer perspective views.
Experimental results show that \method achieves better visual quality, text alignment, and geometric consistency than existing training-free methods while achieving a 15.36$\times$ speedup over the best-performing baseline.

\section{Related work}
\subsection{Diffusion and flow-matching models for image generation}
Diffusion models\cite{Ho2020DDPM} and flow-matching models\cite{liu2023flow,lipman2023flow} have become the dominant backbone for text-to-image generation~\cite{Rombach2022LDM,nichol2022Glide,saharia2022Imagen,Esser2024SD3,flux2024}. 
However, these models are typically trained for a fixed aspect ratio or a narrow range of resolutions, and generation outside that regime often degrades layout and composition unless additional design choices are introduced.

To support flexible output sizes and aspect ratios, prior work has either adapted pre-trained text-to-image diffusion models at inference time~\cite {jin2023trainingfree} or modified the architecture and training framework to natively support variable resolutions and aspect ratios~\cite{wang2025nativeresolution}. 
MultiDiffusion~\cite{Bar-Tal2023MultiDiffusion} and its variants~\cite{lee2023syncdiffusion,wang2024stitchdiffusion,Frolov2025SpotDiffusion,zhang2025multiscalediffusion} extend pre-trained text-to-image diffusion models to larger canvases through region-wise denoising and fusion, enabling arbitrary aspect ratios and higher resolutions in a model-agnostic, zero-shot manner.
However, these methods are primarily designed for planar imagery or wide-canvas images rather than geometrically consistent \thsi panoramas and do not explicitly enforce loop consistency over \thsi scenes.

\subsection{Training-based text-conditional \thsi panorama generation}
Text2Light~\cite{chen2022text2light} pioneered text-driven panorama generation using a discrete VQVAE latent space and an autoregressive prior.
Following the success of diffusion models in text-to-image (T2I) generation, diffusion-based methods have also become the dominant paradigm for panorama generation~\cite{tang2023mvdiffusion,Ye2024DiffPano,panfusion2024,kalischek2025cubediff,ni2025UniPano}.
These methods rely on text-panorama paired datasets, which are challenging to collect at scale and, therefore, tend to have limited domain coverage and reduced robustness to diverse text prompts.
For example, a model trained primarily on indoor panorama datasets typically generalizes poorly to outdoor scenes.
CurvedDiffusion~\cite{Andrey2024CurvedDiff} finetunes a pre-trained T2I latent diffusion model using synthetically distorted images and demonstrates that panorama generation can emerge from such adaptation.
However, the output aspect ratio is inherited from the base model, and it supports only a fixed resolution.

In contrast, our method directly uses pre-trained T2I models with their parameters fixed, thereby inheriting the base models' prompt coverage.
Moreover, it supports arbitrary aspect ratios and resolutions by adopting region-wise generation as in MultiDiffusion-style methods.

\subsection{Training-free text-conditional \thsi panorama generation}
A few prior works have explored training-free panorama generation. 
DynamicScaler~\cite{Liu2025DynamicScaler} optimizes panorama-domain noisy latents over diffusion timesteps by applying MultiDiffusion with equirectangular projection (ERP).
SphereDiff~\cite{park2025spherediff} similarly adopts a training-free formulation, but represents latents directly on the sphere to reduce the distortion near the poles.
Because the original MultiDiffusion formulation assumes one-to-one pixel correspondence between the large canvas and each local patch, these methods must use direct pixel sampling between panoramas to perspective views, such as nearest-neighbor sampling in DynamicScaler and the dynamic latent sampling specialized for the spherical latent in SphereDiff.
To cover the full panorama and maintain seamlessness, they require many overlapping camera views.
As a result, they suffer from inefficient inference due to many neural function evaluations.

An orthogonal line of work studies training-free text-to-3D scene generation.
WonderJouney~\cite{Yu2024WonderJourney}, LucidDreamer~\cite{Chung2025LucidDreamer}, and WonderWorld~\cite{Yu2025WonderWorld} iteratively expand 3D scenes by optimizing 3D point clouds, 3D Gaussian splatting (3DGS)~\cite{Kerbl20233DGS}, and 3D Gaussian surfels, respectively, from inpainted views generated by a pre-trained text-to-image generator.
Because these methods progressively construct the scene through sequential view expansion and optimization, they often suffer from loop inconsistency and error accumulation.
To mitigate these issues, DreamScene360~\cite{zhou2024dreamscene360} first generates several \thsi panoramas followed by VLM-based initial scene selection, and then lifts them into 3D representations.
Our work focuses on \thsi panorama generation and is complementary to these training-free text-to-3D scene generation methods, as exemplified in DreamScene360.

\section{Method}
In this section, we first briefly review MultiDiffusion and discuss the challenges of applying it to \thsi panoramic view synthesis.
We then introduce \method, an extension of MultiDiffusion that supports arbitrary linear projections between the target and reference image spaces.
Since most diffusion models are latent diffusion models~\cite{Rombach2022LDM}, we use ``image'' and ``pixel'' to also refer to a latent tensor and its spatial components, respectively.

\subsection{Preliminaries}

\subsubsection{Overview of MultiDiffusion.}
MultiDiffusion~\cite{Bar-Tal2023MultiDiffusion} enables the generation of images with arbitrary shapes using a pre-trained diffusion model without additional training or finetuning.
Let $\Phi: \prsp \times \condprsp \rightarrow \prsp$ denote a pre-trained diffusion model, where $\prsp \subseteq \R^{M}$ is the reference image space, and $\condprsp$ is the corresponding condition space.
Starting from $I_T \sim \mathcal{N}(\mathbf{0}, \mathbf{I})$ with condition $y \in \condprsp$, the reverse diffusion process is written as
\begin{equation}
    I_T, I_{T-1}, \dots, I_0 
    \quad \text{s.t.} \quad 
    I_{t-1} = \Phi(I_t \mid y),
\end{equation}
which gradually transforms a noisy image $I_T$ into a clean image $I_0$.
MultiDiffusion defines a reverse diffusion process in a potentially different target image space $\pano \subseteq \R^{M'}$ with condition space $\condpano$.
Starting from $J_T \sim \mathcal{N}(\mathbf{0}, \mathbf{I})$ with condition $z \in \condpano$, the reverse MultiDiffusion process is written as
\begin{equation}
    J_T, J_{T-1},\dots, J_0
    \quad \text{s.t.} \quad 
    J_{t-1} = \Psi(J_t  \mid z),
\end{equation}
where $\Psi: \pano \times \condpano \rightarrow \pano$ denotes a denoising operator at diffusion step $t$, referred to as the MultiDiffuser.
In panoramic generation, the target image typically has a larger spatial extent than the reference image, \ie, $M' \geq M$.

The key idea of MultiDiffusion is to construct $\Psi$ so that it is as consistent as possible with the pre-trained diffusion model $\Phi$.
Let $F_{t, i}: \pano \rightarrow \prsp$ denote a mapping from the target image space to the reference image space, where $F_{t, i}(J)$ extracts a reference image from $J$ by direct pixel sampling, such as cropping or nearest neighbor reprojection.
Let $y_{t, i} = \lambda_{t, i}(z)$ be the corresponding condition for the $i$-th reference image, where $\lambda_{t, i}: \condpano \rightarrow \condprsp$ maps the target condition to the reference condition (\eg, LLM-based conversion from global scene condition to that of each local patch).
Then, the MultiDiffuser is defined as
\begin{equation}
\label{eq:multi_diffuser}
    \Psi(J_t \mid z) 
        = \argmin_{J_{t-1} \in \pano} 
            \sum_{i=1}^{N} 
                \left\| 
                    W_{t,i} \odot 
                    \left[ 
                        F_{t,i}(J_{t-1}) - \Phi(F_{t,i}(J_t) \mid y_{t, i}) 
                    \right] 
                \right\|^2,
\end{equation}
where $W_{t, i} \in \R^{M}_{\ge 0}$ is a per-pixel weight and $\odot$ is the Hadamard product.
When $F_{t,i}$ is a direct pixel-sampling operator, \cref{eq:multi_diffuser} admits the closed-form solution
\begin{equation}
    \label{eq:multi_diffuser_closed}
    \Psi(J_t \mid z) 
        = \sum_{i=1}^N   
            \frac{F_{t,i}^{-1}(W_{t,i})}{\sum_{j=1}^N F_{t,j}^{-1}(W_{t,j})} 
            \odot 
            F_{t,i}^{-1}\!\left(\Phi(F_{t,i}(J_t) \mid y_{t, i}) \right).
\end{equation}
In practice, $W_{t, i}$ is often set to $\mathbf{1}_{M} \in \mathbb{R}^M$ so that all reference images contribute equally.
We follow this setting and omit the weight term in the following sections for brevity.

\subsubsection{Challenges of applying MultiDiffusion to \thsi panoramas.}

\begin{figure}[t]
  \centering
  \setlength{\fboxsep}{0pt}    
  \setlength{\fboxrule}{0.5pt} 

  \begin{subfigure}[t]{0.48\linewidth}
    \centering
    \includegraphics[width=\linewidth]{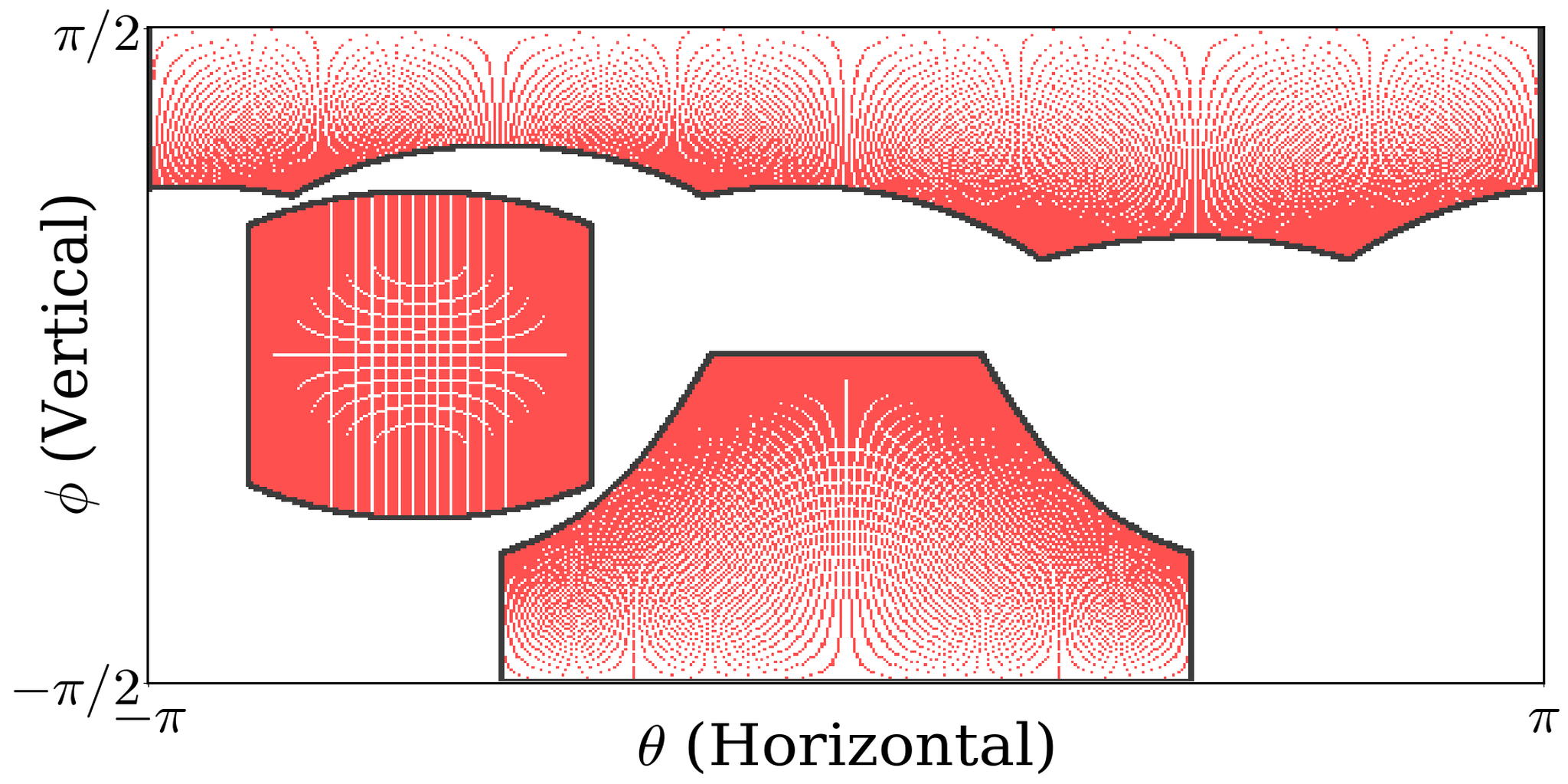}
    \caption{ERP + direct pixel sampling from perspective views to panorama.}
    \label{fig:issue_of_direct_pixel_sampling_a}
  \end{subfigure}
  \hfill
  \begin{subfigure}[t]{0.48\linewidth}
    \centering
    \includegraphics[width=\linewidth]{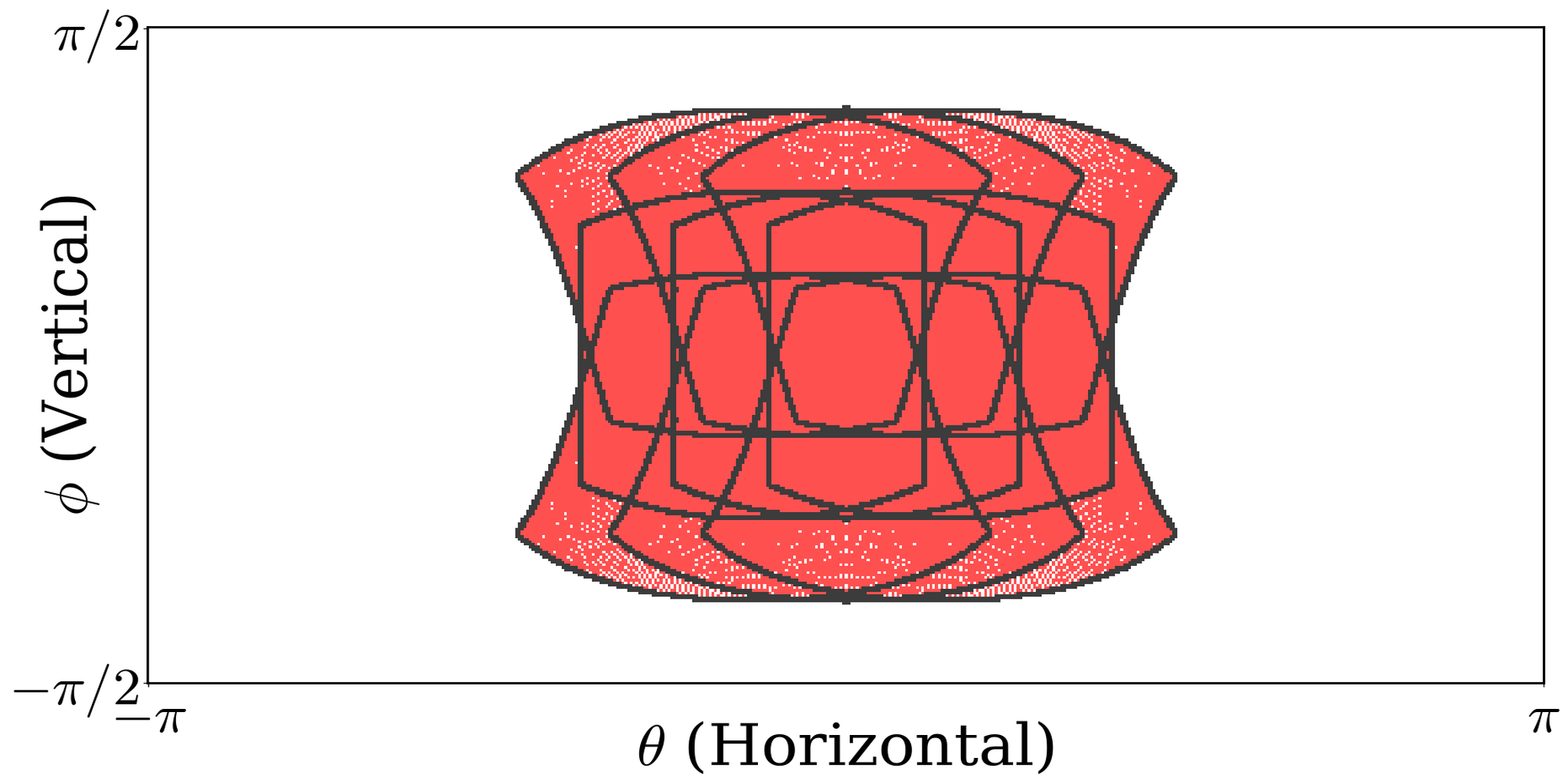}
    \caption{Covered pixels by nine camera positions near the center used in SphereDiff~\cite{park2025spherediff}.}
    \label{fig:issue_of_direct_pixel_sampling_b}
  \end{subfigure}

  \caption{Perspective-to-panorama mapping with direct pixel sampling. Panorama size: $256\times512$; perspective view: $128\times128$ with ${\rm FOV}=\ang{90}$. (a) Pixels covered by a single view projected onto the ERP panorama at different camera latitudes $\phi=[\ang{77.5},\ang{0},\ang{-45}]$ (top to bottom). ERP-based direct sampling leaves many pixels uncovered (white pixels within the view frustum). (b) Existing methods mitigate this issue by densely sampling overlapping cameras, but they must run the diffusion model for every view, increasing computational cost.}
  \label{fig:issue_of_direct_pixel_sampling}
\end{figure}

Applying \cref{eq:multi_diffuser_closed} to non-perspective \thsi panoramas is challenging because the direct-sampling assumption makes it difficult to cover all panorama pixels.
To use \cref{eq:multi_diffuser_closed}, every target pixel must be assigned to at least one pixel in some perspective view.
However, when perspective views are projected onto an ERP panorama and then reprojected by direct pixel sampling through $F_{t, i}^{-1}$, many panorama pixels remain uncovered, as illustrated in \cref{fig:issue_of_direct_pixel_sampling_a}.
Existing approaches~\cite{Liu2025DynamicScaler,park2025spherediff} address this issue by densely sampling camera views with substantial overlap (\cref{fig:issue_of_direct_pixel_sampling_b}).
Since the pre-trained diffusion model $\Phi$ must be evaluated for every perspective view at every diffusion step, this dense sampling leads to high computational cost (\eg, SphereDiff requires more than 30 minutes per panorama, as shown in \cref{tab:baseline_comparison}).

\begin{figure}[t]
  \centering

  \begin{subfigure}[c]{0.32\linewidth}
    \centering
    \begin{minipage}[c][3cm][c]{\linewidth}
      \centering
      \includegraphics[width=0.95\linewidth,keepaspectratio]{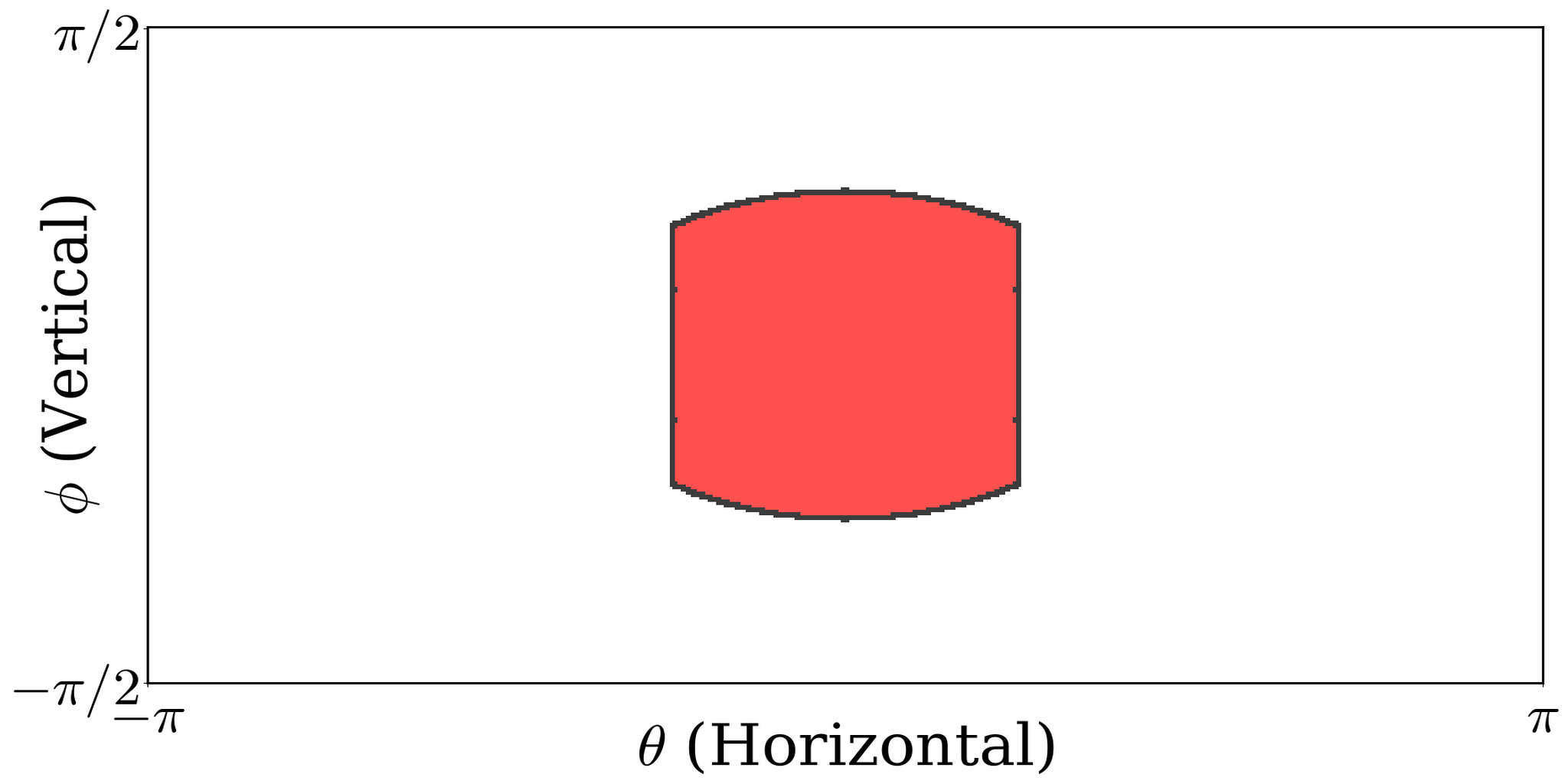}
    \end{minipage}
    \caption{Bilinear mapping}
    \label{fig:issue_of_bilinear_plus_multidiffusion_a}
  \end{subfigure}
  \hfill
  \begin{subfigure}[c]{0.32\linewidth}
    \centering
    \begin{minipage}[c][3cm][c]{\linewidth}
      \centering
      \includegraphics[width=0.95\linewidth,keepaspectratio]{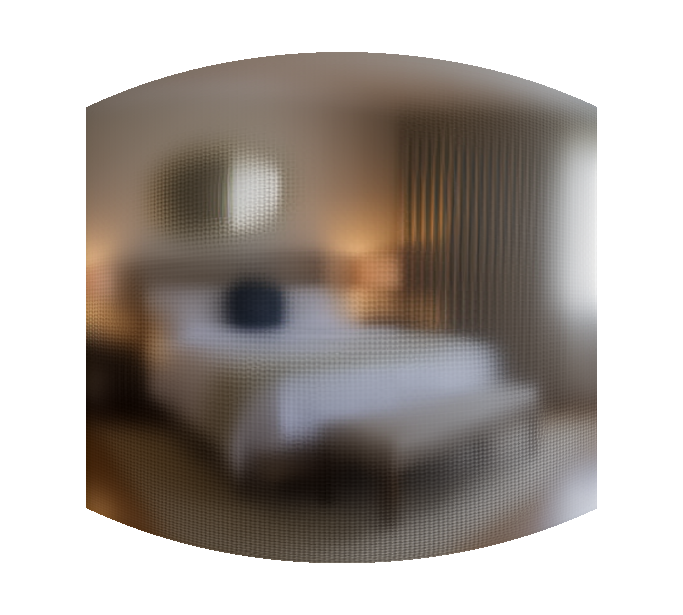}
    \end{minipage}
    \caption{Naive extension of \cref{eq:multi_diffuser_closed}}
    \label{fig:issue_of_bilinear_plus_multidiffusion_b}
  \end{subfigure}
  \hfill
  \begin{subfigure}[c]{0.32\linewidth}
    \centering
    \begin{minipage}[c][3cm][c]{\linewidth}
      \centering
      \includegraphics[width=0.95\linewidth,keepaspectratio]{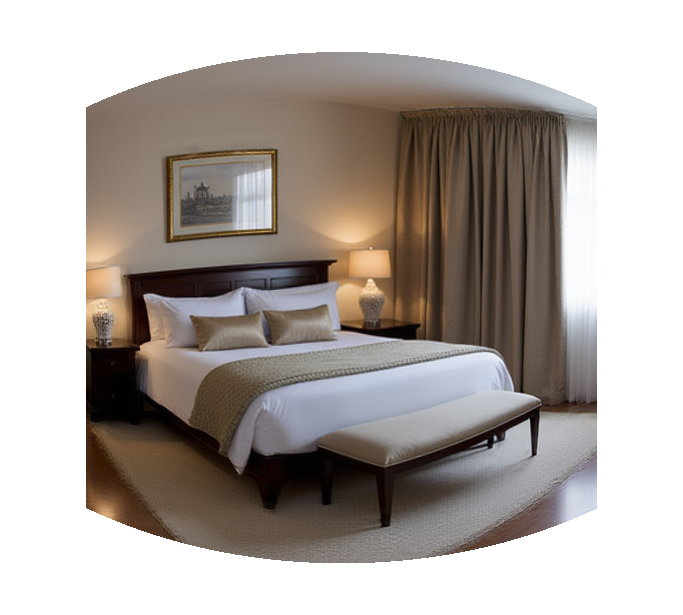}
    \end{minipage}
    \caption{Ours}
    \label{fig:issue_of_bilinear_plus_multidiffusion_c}
  \end{subfigure}

  \caption{Panorama generation with bilinear mapping between the panorama and a perspective view, where the panoramic latent is optimized at each denoising step. We use \textsc{FLUX}~\cite{flux2024} as the pre-trained diffusion model with the text prompt \textit{"A room"}. (a) A perspective image is bilinearly mapped to the center of the panorama, leaving no pixels uncovered. (b) Naively extending \cref{eq:multi_diffuser_closed} by pixel-wise averaging produces a blurry result. (c) Our method generates a high-quality image under the same bilinear mapping.}
  \label{fig:issue_of_bilinear_plus_multidiffusion}
\end{figure}

A natural way to reduce the number of uncovered pixels is to use denser mappings, such as bilinear interpolation or pixel splatting from panoramas to perspective views.
As shown in \Cref{fig:issue_of_bilinear_plus_multidiffusion_a}, bilinear mapping between a perspective image and the center of a panorama produces no uncovered pixels.
One may therefore consider extending \cref{eq:multi_diffuser_closed} by still averaging reprojected predictions pixel-wise under such dense mappings.
However, this naive extension violates consistency with the pre-trained model and severely degrades the intermediate denoising latents.
To demonstrate this, we generate only the center region of a panorama using MultiDiffusion with bilinear mapping between the perspective image and the panorama.
As shown in \cref{fig:issue_of_bilinear_plus_multidiffusion_b}, the naive extension produces noticeably degraded results.

\subsection{MultiDiffuser with arbitrary linear projections}
To overcome this limitation, we propose \methodfull (\method), an extension of MultiDiffusion to the case where the target-to-reference mapping is an arbitrary linear operator.
Let $\mapset \subset \mathbb{R}^{M\times M'}$ denote a family of linear mappings from a target panorama vector
$J \in \pano \subseteq \mathbb{R}^{M'}$ to a reference image vector in $\R^{M}$.
We replace the direct-sampling operators $\{F_{t,i}\}_{i=1}^{N}$ with linear projection matrices $\{S_{t,i} \in \mapset \}_{i=1}^{N}$.
Then, the resulting linear-fusion MultiDiffuser $\Psi_{\mathrm{LF}}$ is defined as
\begin{equation}
    \label{eq:psi_lf_def}
    \Psi_{\rm LF}(J_t \mid z) = 
        \argmin_{J_{t-1}\in\pano} 
            \sum_{i=1}^{N} 
                \left\| 
                    S_{t,i}J_{t-1} - \Phi\!\left(S_{t,i}J_t \mid y_{t, i}\right) 
                \right\|^2.
\end{equation}
Here, $S_{t, i}J \in \mathbb{R}^{M}$ denotes the $i$-th reference image obtained from $J$ by the corresponding linear projection.

\subsubsection{Regularized formulation.}
Solving \cref{eq:psi_lf_def} in closed form would require explicitly forming and inverting the associated normal-equation matrix, which is computationally prohibitive for high-resolution panoramas.
Instead, we introduce a quadratic regularizer and estimate $J_{t-1}$ by solving the following regularized least-squares problem:
\begin{equation}
    \label{eq:reg_lsq}
    J_{t-1} =
        \argmin_{J\in\pano}
            \sum_{i=1}^{N}
                \left\| 
                    S_{t,i}J - \Phi\!\left(S_{t,i}J_t \mid y_{t, i}\right) 
                \right\|^2 
                + \lambda \|LJ\|^2,
\end{equation}
where $\lambda \geq 0$ is a regularization coefficient, and $L=W^{1/2}D \in \mathbb{R}^{P\times M'}$ is a weighted first-order regularizer with a discrete difference operator $D \in \mathbb{R}^{P\times M'}$ and a diagonal weight matrix $W=\mathrm{diag}(\mathbf{w}) \in \mathbb{R}^{P\times P}$ for $\mathbf{w} \in \mathbb{R}^P_{\ge 0}$.
This formulation includes, for example, ridge- or Laplacian-type regularization depending on the choice of $D$ and $W$.

\subsubsection{Augmented system.}
We rewrite \cref{eq:reg_lsq} in an equivalent augmented form.
Define the stacked operator and target vector as
\begin{equation}
    \label{eq:augmented_system}
    \tilde S := 
        \begin{bmatrix}
            S_{t,1}\\
            \vdots\\
            S_{t,N}\\
            \sqrt{\lambda}\,L
        \end{bmatrix}
        \in \R^{(NM+P)\times M'},
    \qquad
    \tilde d :=
        \begin{bmatrix}
            \Phi(S_{t, 1}J_t \mid y_{t, 1})\\
            \vdots\\
            \Phi(S_{t,N}J_t \mid y_{t, N})\\
            \mathbf{0}
        \end{bmatrix}
        \in \R^{NM+P}.
\end{equation}
Then, \cref{eq:reg_lsq} is equivalently written as
\begin{equation}
    \label{eq:augmented_lsq}
    J_{t-1} = 
        \argmin_{J \in \pano}\ 
        \|\tilde S J - \tilde d\|^2.
\end{equation}
We solve \cref{eq:augmented_lsq} using a matrix-free Krylov subspace method, either by applying preconditioned conjugate gradients (PCG)~\cite{Hestenes1952CG} to the corresponding normal equations or by directly applying LSMR~\cite{Fong2011LSMR}.
In all cases, the solver requires only the operations $J \mapsto S_{t, i}J$, $I \mapsto S_{t, i}^{\top} I$, and $J \mapsto L^{\top}(LJ)$, together with their stacked counterparts for $\tilde S$.
Therefore, the projection matrices need not be formed explicitly in memory during the iterative update.

\begin{figure}[t]
    \centering
    \includegraphics[width=\linewidth]{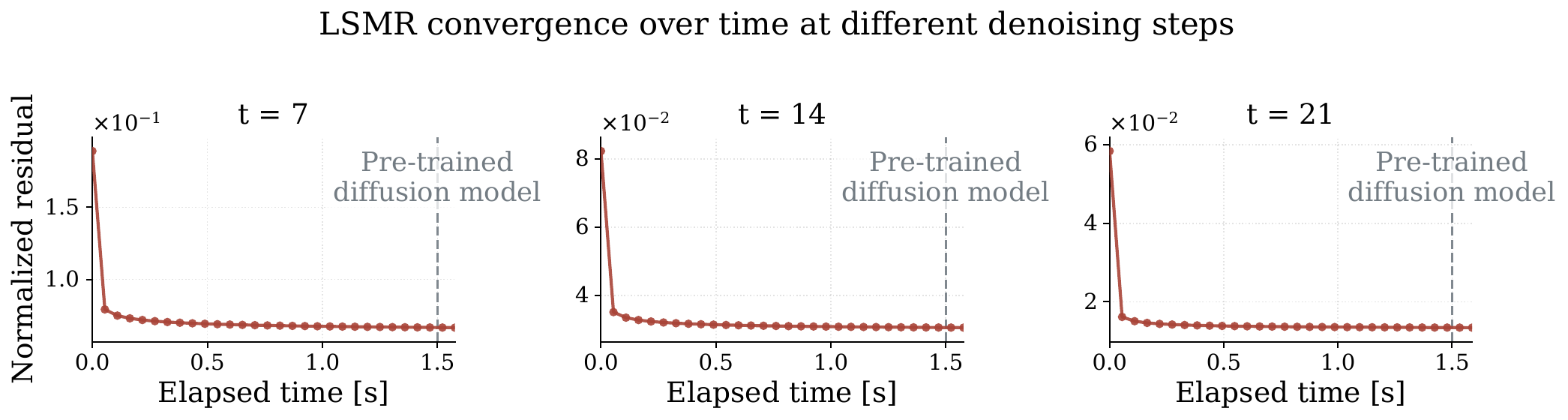}
    \caption{Runtime comparison between the LSMR~\cite{Fong2011LSMR} update and the total denoising cost of a full MultiDiffusion step. We plot the normalized residual in \cref{eq:psi_lf_def} every 10 solver updates at denoising timestep $t \in \{7, 14, 21\}$. The total denoising cost for $N=14$ using Flux~\cite{flux2024} is shown as a gray dashed line. The Krylov solver update converges far more quickly than the denoising cost of the corresponding MultiDiffusion step.}
    \label{fig:iterative_update_runtime}
\end{figure}

\Cref{fig:iterative_update_runtime} compares the runtime of the LSMR update with the total denoising cost of one MultiDiffusion step at different denoising steps.
For this experiment, we used $N = 14$, which is also the default setting in our main experiments.
We plot the normalized residual $\sum_i \| S_{t,i}J - \Phi(S_{t,i}J_t \mid y_{t, i}) \|^2 \big/ \sum_i\| \Phi(S_{t,i}J_t \mid y_{t, i})\|^2$ every 10 solver iterations.
The residual decreases rapidly within a few tens of iterations, and the solver update requires less than 20\% of the computational cost of a full MultiDiffusion step.
This overhead is substantially smaller than the additional cost incurred by increasing the number of views $N$, which scales linearly with the number of diffusion model evaluations. 
For reference, DynamicScaler and SphereDiff use $N=44$ and $N=89$, respectively, corresponding to roughly 3-7$\times$ longer denoising runtime.
These results show that the proposed iterative update adapts the MultiDiffusion to panoramic view synthesis in a computationally efficient manner.

\subsection{Implementation of the target space and projection}

\subsubsection{Definition of the reference image space.}
We define the reference image space $\prsp$ as a rectangle of size $H \times W$.
The mapping operations $J \mapsto S_{t,i}J$ and $I \mapsto S_{t,i}^{\top} I$ are implemented using pixel correspondences in 3D space.
To establish these correspondences, we assume that each pixel in the reference image corresponds to a point on the unit sphere.
Specifically, for a pixel $\mathbf{u} \in [0, W) \times [0, H)$ in $I_{t,i} \in \prsp$, we define the corresponding 3D point $\mathbf{x} \in \mathbb{S}^2$ by back-projecting its viewing ray onto the unit sphere, where $\mathbf{u}$ and $\mathbf{x}$ satisfy
\begin{equation}
    \label{eq:camera_projection}
    \mathbf{u} \sim M_{t,i} [\mathbf{x}^\top, 1]^\top,
    \quad
    M_{t,i} = K [R_{t,i} \mid \mathbf{c}],
\end{equation}
where $M_{t,i}$ denotes the camera projection matrix, $K \in \mathbb{R}^{3 \times 3}$ is the intrinsic matrix, $R_{t,i} \in \mathbb{R}^{3 \times 3}$ is the rotation matrix, and $\mathbf{c} \in \mathbb{R}^{3}$ is the translation vector.

The intrinsic matrix $K$ and the translation vector $\mathbf{c}$ are shared across all views and timesteps.
Unless otherwise noted, we set ${\rm FOV}_x = {\rm FOV}_y = \ang{90}$ when computing $K$, and use $\mathbf{c}=[0,0,0]^\top$ since all camera centers coincide with the origin.
In contrast, the rotation matrix $R_{t,i}$ depends on both the timestep and the view index.
Following the Panoramic Projecting Denoising proposed in DynamicScaler~\cite{Liu2025DynamicScaler}, we sweep the viewing direction horizontally as the reverse diffusion process proceeds, allowing the model to efficiently extract different subregions of the target image space over time.
Specifically, let
$\{(\theta_i, \phi_i)\}_{i=1}^{N}$ denote the initial longitude-latitude pairs with $\theta_i \in (-\pi,\pi]$ and $\phi_i \in (-\pi/2,\pi/2]$.
At timestep $t$, we define the view angle as
$(\theta_{t,i}, \phi_{t,i}) = (\theta_i + st, \phi_i)$,
where $s \in (-\pi,\pi)$ is a constant angular shift per timestep.
We then construct the rotation matrix $R_{t,i}(\theta_{t,i}, \phi_{t,i})$ such that its forward axis is aligned with the viewing direction $d(\theta_{t,i},\phi_{t,i})$ defined by $d(\theta, \phi) = (\cos \phi \sin \theta, \, \sin \phi, \, \cos \phi \cos \theta)$.

\subsubsection{ERP latent space with bilinear projection.}
Inspired by DynamicScaler~\cite{Liu2025DynamicScaler}, we represent the target space $\pano$ as an ERP latent.
Specifically, we define $\pano$ as a rectangle of size $H_p \times W_p$, where each pixel $(i, j) \in [0, W_p) \times [0, H_p)$ corresponds to a point on the sphere with longitude $\theta$ and latitude $\phi$ as
\begin{equation}
 \label{eq:erp_pixel_to_lonlat}
  \theta = 2\pi \cdot \frac{i}{W_p} - \pi,
  \quad
  \phi = \pi \cdot \frac{j}{H_p} - \frac{\pi}{2}.
\end{equation}

We define the mapping $J \mapsto S_{t,i}J$ from the panorama to a perspective view by bilinear interpolation.
Specifically, to compute the pixel value $I[u, v]$ at pixel position $\mathbf{u} = (u, v)$, we first back-project its viewing ray to a 3D point $(x_{\mathbf{u}}, y_{\mathbf{u}}, z_{\mathbf{u}})$ using \cref{eq:camera_projection}, and then compute the corresponding longitude and latitude $(\theta_{\mathbf{u}}, \phi_{\mathbf{u}}) \in \mathbb{R}^2$ as $\theta_{\mathbf{u}} = \operatorname{atan2}(x_{\mathbf{u}}, z_{\mathbf{u}})$ and $\phi_{\mathbf{u}} = \arcsin(y_{\mathbf{u}})$.
The corresponding continuous panorama coordinate $(i_{\mathbf{u}}, j_{\mathbf{u}})$ is then obtained by inverting \cref{eq:erp_pixel_to_lonlat}.
Using the 1D linear interpolation kernel $\kappa(r) = \max(0, 1 - |r|)$, we compute $I[u, v]$ as
\begin{equation}
    \label{eq:bilinear_gather}
    I[u, v]
    =
    \sum_{(i, j) \in \mathcal{N}(i_{\mathbf{u}}, j_{\mathbf{u}})}
    \kappa(i - i_{\mathbf{u}})
    \kappa(j - j_{\mathbf{u}})
    J[i, j],
\end{equation}
where $\mathcal{N}(i_{\mathbf{u}}, j_{\mathbf{u}}) = \{(\floor{i_{\mathbf{u}}} + \delta_i,\; \floor{j_{\mathbf{u}}} + \delta_j) \mid \delta_i, \delta_j \in \{0, 1\}\}$ denotes the four neighboring panorama pixels of $(i_{\mathbf{u}}, j_{\mathbf{u}})$.
The horizontal index is treated as periodic, whereas the vertical index is clipped to the valid range.
For the reverse mapping $I \mapsto S_{t,i}^{\top} I$, we use the same correspondences and interpolation weights as in \cref{eq:bilinear_gather}, and scatter each pixel value $I[u, v]$ onto the target ERP latent $J$.

\begin{algorithm}[t]
\caption{Panorama generation with \method}
\label{alg:linear_mapping_multi_diffusion}
\begin{algorithmic}[1]
\Require Pre-trained diffusion model $\Phi$, view conditions $\{y_{t,i}\}_{i=1}^N$, linear mappings $\{S_{t,i}\}_{i=1}^{N}$, number of views $N$, total number of diffusion steps $T$, and \method stopping timestep $T_M$
\State $J_T \sim \mathcal{N}(\mathbf{0}, \mathbf{I})$

\Statex \textbf{Phase 1: Panorama latent optimization (\method)}
\For{$t = T, T-1, \dots, T_M + 1$}
    \State $I_{t,i} \leftarrow S_{t,i}J_t \quad \forall i \in [N]$
    \Comment{Render}
    
    \State $\tilde{I}_{t-1,i} \leftarrow \Phi(I_{t,i}, t, y_{t,i}) \quad \forall i \in [N]$
    \Comment{Denoise}

    \State $J_{t-1} \leftarrow \mathrm{KrylovSolve}\!\left(\{S_{t,i}\}_{i=1}^{N}, \{\tilde{I}_{t-1,i}\}_{i=1}^{N}\right)$
    \Comment{Solve \cref{eq:augmented_lsq}}
\EndFor

\Statex \textbf{Phase 2: View-wise post refinement}
\State $I_{T_M,i} \leftarrow S_{T_M,i}J_{T_M} \quad \forall i \in [N]$
\Comment{Render}

\For{$t = T_M, T_M-1, \dots, 1$}
    \State $I_{t-1,i} \leftarrow \Phi(I_{t,i}, t, y_{t,i}) \quad \forall i \in [N]$
    \Comment{Denoise}
\EndFor

\State $J_0 \leftarrow \mathrm{DistortionAwareWeightedAveraging}\!\left(\{S_{T_M,i}\}_{i=1}^{N}, \{I_{0,i}\}_{i=1}^{N}\right)$
\Comment{Aggregate}
\State \Return $J_0$
\end{algorithmic}
\end{algorithm}

\subsection{View-wise post refinement}
Although \method produces globally consistent panorama latents, the final decoded ERP images can still appear slightly blurry compared with images generated directly by the base model in its native perspective domain.
We conjecture that this gap arises because the panorama is optimized in the VAE~\cite{Kingma2014VAE} latent space and decoded into the ERP format, where both latent-space compression and the geometric distortion inherent to ERP representations can weaken local details.

In practice, we stop the panorama optimization at an intermediate timestep $T_M > 0$ and use the resulting latent $J_{T_M}$ for a view-wise post-refinement stage.
Specifically, we first render a set of perspective-view latents $\{I_i\}_{i=1}^{N_{\rm ref}}$ by applying the corresponding projections $S'_{T_M,i}$ to $J_{T_M}$.
We then continue denoising each rendered view independently using the same pre-trained image generator $\Phi$ until the final step.
Intuitively, this step brings each local view back to the native generation domain, allowing high-frequency details to be recovered while preserving the global structure established by \method.

After denoising, each refined latent is decoded independently into an RGB image.
We then project the refined perspective images back to the ERP image and aggregate them using Distortion-Aware Weighted Averaging from SphereDiff~\cite{park2025spherediff}, which accounts for the non-uniform distortion of ERP coordinates.
\Cref{alg:linear_mapping_multi_diffusion} summarizes an overview of the proposed generation process.

\section{Experiments}

\subsection{Experimental setup}
\subsubsection{Implementation and inference details.}
We used \textsc{FLUX}~\cite{flux2024} as the pre-trained image generator with the classifier-free guidance~\cite{ho2021CFG} scale set to 3.5.
For text prompts, we followed SphereDiff~\cite{park2025spherediff} and used their 20 prompt sets. 
For each prompt set, we generated 10 panorama images with different random seeds.
The total number of denoising steps $T$ is set to 28, and the stopping step for \method $T_M$ is set to 23.

To construct mappings between the panorama and perspective views, we used $N=14$ initial view directions: two polar views $(\theta, \phi) = (0, \pm\frac{\pi}{2})$, eight tilted views $(\theta, \phi) = (\frac{\pi}{2}i, \pm\frac{\pi}{3})$ for $i \in \{0, 1, 2, 3\}$, and four equatorial views $(\theta, \phi) = (\frac{\pi}{2}i, 0)$ for $i \in \{0, 1, 2, 3\}$.
We set the per-step angular shift $s = \frac{\pi}{18}$.
We generated $2048 \times 4096$ ERP panoramas with $512 \times 512$ perspective views.
All experiments were conducted on a single NVIDIA A100 GPU (40GB).

\subsubsection{Baselines.}
We compared against DynamicScaler~\cite{Liu2025DynamicScaler} and SphereDiff~\cite{park2025spherediff} as training-free baselines.
We set $N=44$ for DynamicScaler and $N=89$ for SphereDiff, which are the default values.
For a fair comparison, we used \textsc{FLUX}~\cite{flux2024} as the base text-to-image generator.
We additionally reported results for representative training-based models, Text2Light~\cite{chen2022text2light} and PanFusion~\cite{panfusion2024}.

\subsubsection{Evaluation process and metrics.}
We followed the evaluation protocol of SphereDiff~\cite{park2025spherediff}.
Specifically, we rendered 14 perspective views from each generated panorama using ${\rm FOV} = \ang{90}$ for evaluation.
Inspired by VBench~\cite{huang2023vbench}, we used MUSIQ~\cite{Ke2021MUSIQ} to assess imaging quality (e.g., exposure, noise, and blur), the LAION aesthetic predictor~\cite{LAION2022AestheticPredictor} to measure aesthetic preference, and CLIP~\cite{radford2021CLIP} feature similarity between each view and the corresponding text prompt to evaluate text alignment.
We also reported Q-Align~\cite{wu2023qalign}, an LLM-based visual evaluator, as an additional measure of perceptual quality.
Furthermore, we conducted the VLM-based evaluation~\cite{Fei2024VLMEval} similar to SphereDiff~\cite{park2025spherediff} to assess the panoramic distortion and continuity using Qwen-VL~\cite{bai2025qwen25vltechnicalreport}.
We use the text prompt proposed in SphereDiff~\cite{park2025spherediff} with slight modifications tailored for Qwen-VL (see Section C in the supplementary material for the full evaluation prompt).
For runtime, we measured the end-to-end generation time excluding I/O overhead.

\subsubsection{Additional analyses in the supplementary material.}
Beyond the main experimental results presented in this section, we provide additional analyses in the supplementary material. 
They include experiments with SANA~\cite{xie2025sana}, an extension to text-to-panoramic video generation with LTX-Video~\cite{HaCohen2024LTXVideo}, latent and projection variants, a sensitivity study on the post-refinement timestep, and additional generated samples.

\subsection{Comparison with baselines}

\subsubsection{Quantitative comparison.}

\begin{table*}[t]
\small
    \centering
    \caption{Comparison with baselines. 2048 $\times$ 4096 panoramas are generated except for PanFusion (which only supports 512 $\times$ 1024). For the training-free methods, Flux~\cite{flux2024} is used as a base T2I model. Runtime denotes the mean wall-clock time measured on a single Nvidia A100 40GB GPU (lower is better). Higher is better for all other metrics. \method supports a dense mapping between panoramic and perspective views, thereby achieving higher generation quality with faster inference. *: The runtime for PanFusion is shown for reference due to its small resolution.}
    \begin{NiceTabular}{l@{\hspace{6pt}}c@{\hspace{6pt}}c@{\hspace{6pt}}c@{\hspace{6pt}}c@{\hspace{6pt}}c@{\hspace{6pt}}c@{\hspace{4pt}}c}
    \toprule
    Method & Runtime & Aesthetic & Imaging & QAlign & CLIP & Distortion & Continuity \\
    \midrule
    \multicolumn{8}{l}{\rowcolor{headerColor} {\textbf{Training-based panorama generation}} \hfill} \\
    Text2Light & 56s & 0.424 & 0.434 & 2.22 & 19.60 & 3.26 & 3.56 \\
    PanFusion & 27s$^*$ & 0.473 & 0.527 & 2.62 & 25.44 & 1.93 & 2.31  \\
    \midrule
    \multicolumn{8}{l}{\rowcolor{headerColor} {\textbf{Training-free panorama generation}} \hfill} \\
    DynamicScaler & 7m 31s & 0.463 & 0.516 & 3.11 & 26.32 & 3.15 & 3.83 \\
    SphereDiff & 32m 15s & 0.597 & 0.533 & 3.23 & 27.65 & \textbf{4.64} & 4.82 \\ 
    \rowcolor{highlightColor} Ours & \textbf{2m 6s} & \textbf{0.616} & \textbf{0.568} & \textbf{3.34} & \textbf{29.19} & \textbf{4.64} & \textbf{4.86} \\
    \bottomrule
    \end{NiceTabular}
    \label{tab:baseline_comparison}
\end{table*}
\begin{figure}[t]
    \centering
    \includegraphics[width=\linewidth]{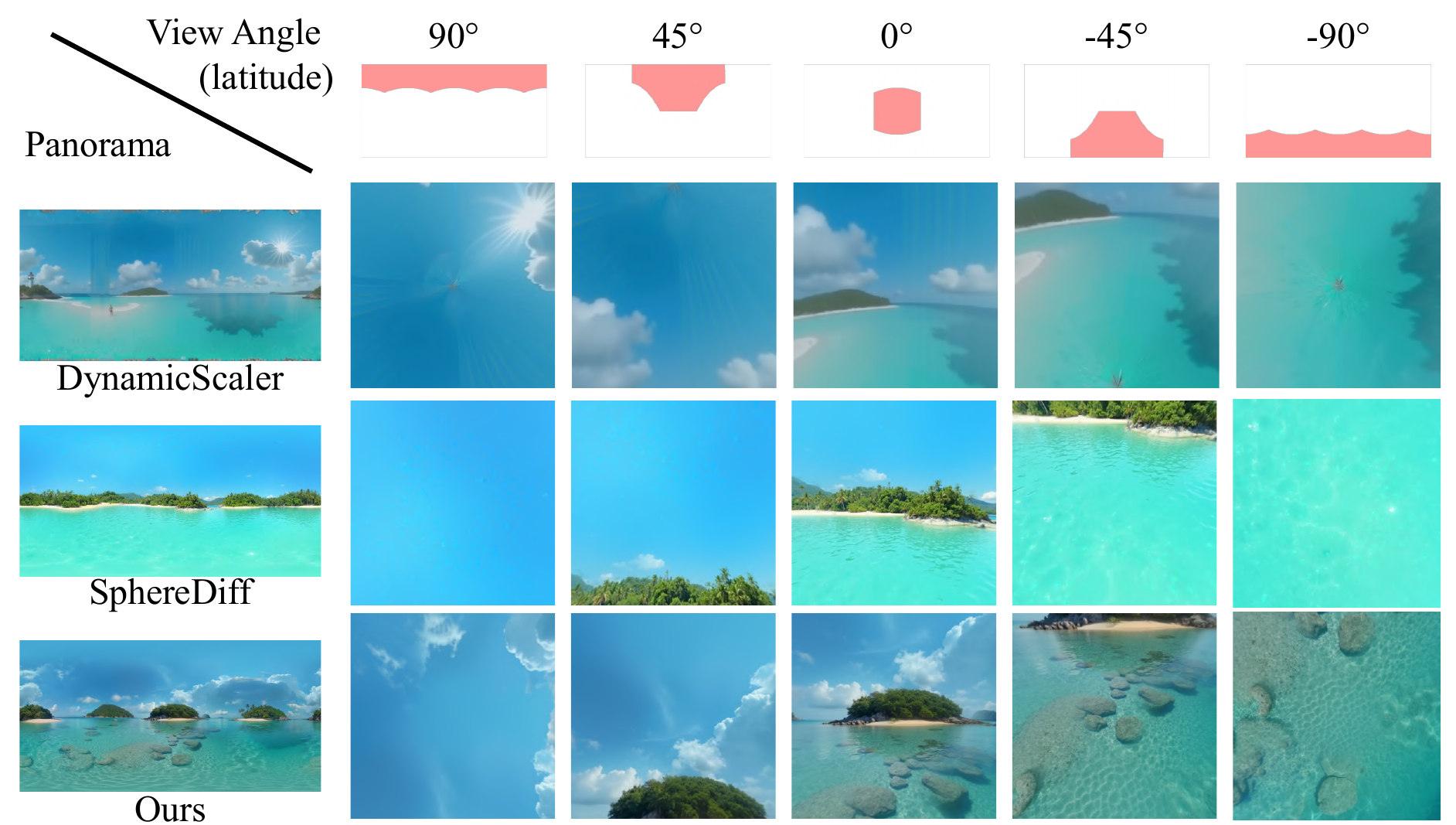}
    \caption{Subjective comparison. DynamicScaler often exhibits pole distortions ($\pm\ang{90}$). SphereDiff and \method generate natural views across directions, and \method achieves higher overall fidelity.}
    \label{fig:subjective_comparison}
\end{figure}
\Cref{tab:baseline_comparison} compares our method with baselines.
Training-based methods often struggle with out-of-domain prompts due to the limited coverage of the training data.
For instance, for the prompt \textit{"Aurora"}, we observed that Text2Light tended to produce a collapsed black image, while PanFusion often generated an unrelated indoor scene.
Among training-free approaches, \method outperforms DynamicScaler and SphereDiff across all metrics.
DynamicScaler applies an additional offset-shifting denoising stage that treats the panorama as a wide-canvas perspective image, rather than enforcing a strictly spherical (ERP-consistent) representation throughout.
This can introduce geometric inconsistencies and degrade quality when evaluated via perspective-view rendering.
In contrast, \method and SphereDiff explicitly model curved panoramic geometry and mitigate these distortions, resulting in consistently higher scores across evaluation metrics.
We attribute the improvement of \method over SphereDiff to the use of denser linear projections, which stabilize optimization on curved representations and enable coherent panorama generation with substantially fewer views.
\Cref{fig:subjective_comparison} subjectively compares the generated panorama and perspective views among training-free approaches.

Notably, \method achieves $3.58\times$ and $15.36\times$ speedup over DynamicScaler and SphereDiff, respectively.
The dominant cost in training-free panorama generation is the number of evaluations of the pre-trained image generator, which scales linearly with the number of perspective views. 
By removing the direct-sampling constraint and supporting dense projections between panoramas and perspective views, \method achieves better quality with significantly lower computational cost.

\subsubsection{User study.}
We further conducted a user study to evaluate the subjective quality of the generated panoramas.
Participants rated samples from DynamicScaler, SphereDiff, and \method in terms of image quality, geometric naturalness, and continuity on a 5-point Likert scale.
As shown in \Cref{tab:user_study}, \method achieved the highest scores across all three aspects, indicating that our formulation improves perceived visual quality while preserving geometric naturalness and seamlessness.
See the supplementary material for the setup details.

\begin{table*}[t]
\small
    \centering
    \caption{User study results on subjective panorama quality. Participants rated each method on a 5-point Likert scale. Higher scores indicate better image quality, fewer perceived distortions, and better continuity. We report the mean rating with 95\% confidence intervals.}
    \begin{NiceTabular}{l@{\hspace{8pt}}c@{\hspace{8pt}}c@{\hspace{8pt}}c}
    \toprule
    Method & Image Quality $\uparrow$ & Distortion $\uparrow$ & Continuity $\uparrow$ \\
    \midrule
    DynamicScaler & 2.39\pmnum{0.17} & 2.36\pmnum{0.16} & 2.25\pmnum{0.20} \\
    SphereDiff & 3.64\pmnum{0.14} & 3.73\pmnum{0.12} & 3.96\pmnum{0.12} \\
    \rowcolor{highlightColor} \method (Ours) & \textbf{4.07}\pmnum{0.10} & \textbf{3.75}\pmnum{0.16} & \textbf{4.07}\pmnum{0.11} \\
    \bottomrule
    \end{NiceTabular}
    \label{tab:user_study}
\end{table*}

\subsection{Ablation studies}
\begin{table}[t]
\small
\centering
\begin{minipage}[t]{0.48\linewidth}
  \centering
  \caption{Results when using different solvers and the number of solver iterations.}
  \label{tab:solver_ablation}
  \begin{NiceTabular}{c@{\hspace{4pt}}c@{\hspace{4pt}}c@{\hspace{4pt}}c}
    \toprule
    Solver & Iters & QAlign$\uparrow$ & CLIP$\uparrow$ \\
    \midrule
    \multirow{3}{*}{PCG}
      & 10  & 3.00 & 28.94 \\
      & 30  & 3.04 & 28.85 \\
      & 100 & 2.97 & 28.95 \\
    \midrule
    \multirow{3}{*}{LSMR}
      & 10  & 3.27 & 29.07 \\
      \rowcolor{highlightColor} & 30  & \textbf{3.34} & \textbf{29.19} \\
      & 100 & 3.30 & 28.93 \\
    \bottomrule
  \end{NiceTabular}
\end{minipage}
\hfill
\begin{minipage}[t]{0.48\linewidth}
  \centering
  \caption{Results when using different regularizers and coefficients.}
  \label{tab:reg_ablation}
  \begin{NiceTabular}{c@{\hspace{4pt}}c@{\hspace{4pt}}c@{\hspace{4pt}}c}
    \toprule
    Regularizer & $\lambda$ & QAlign$\uparrow$ & CLIP$\uparrow$ \\
    \midrule
    - & 0 & 3.282 & 29.02 \\
    \midrule
    \multirow{3}{*}{Ridge}
      & $10^{-3}$ & 3.330 & 29.10 \\
      & $10^{-4}$ & 3.340 & 29.03 \\
      & $10^{-5}$ & 3.336 & 28.99 \\
    \midrule
    \multirow{3}{*}{Laplacian}
      & $10^{-3}$ & 3.297 & 28.89 \\
      \rowcolor{highlightColor} & $10^{-4}$ & \textbf{3.341} & \textbf{29.19} \\
      & $10^{-5}$ & 3.321 & 29.00 \\
    \bottomrule
  \end{NiceTabular}
\end{minipage}
\end{table}

\subsubsection{Solver types and number of updates.}
\Cref{tab:solver_ablation} reports the effect of the solver choice and the number of solver iterations.
We evaluated PCG~\cite{Hestenes1952CG} and LSMR~\cite{Fong2011LSMR} with iteration in $\{10, 30, 100\}$.
For PCG, we used a diagonal preconditioner derived from the squared interpolation weights computed from $S_{i,t}^{\top}S_{i,t}$.
Overall, LSMR achieves better scores across all metrics.
Increasing the number of iterations from 10 to 30 improved performance, whereas further increasing it to 100 yielded diminishing returns.

\subsubsection{Regularizer and its coefficient.}
\Cref{tab:reg_ablation} reports an ablation on regularizer types and the coefficient $\lambda$.
Using LSMR, we compared two regularizers: ridge and discrete Laplacian.
For each regularizer, we tested $\lambda \in \{10^{-3}, 10^{-4}, 10^{-5}\}$.
Across these settings, both regularizers performed comparably.
We therefore used the discrete Laplacian with $\lambda = 10^{-4}$ as the default, as it provided consistently higher \textit{QAlign} and \textit{CLIP} scores.

\subsubsection{Number of reference perspective views.}

\begin{table*}[t]
\small
    \centering
    \caption{Results when using a different number of perspective views $N$.}
    \begin{NiceTabular}{c@{\hspace{4pt}}c@{\hspace{4pt}}c@{\hspace{4pt}}c@{\hspace{4pt}}c@{\hspace{4pt}}c@{\hspace{4pt}}c @{\hspace{4pt}}c}
    \toprule
    $N$ & Runtime$\downarrow$ & Aesthetic$\uparrow$ & Imaging$\uparrow$ & QAlign$\uparrow$ & CLIP$\uparrow$ & Distortion$\uparrow$ & Continuity$\uparrow$ \\
    \midrule
    6 & 1m 15s & 0.518 & \textbf{0.634} & \textbf{3.549} & 27.55 & 3.98 & 4.53 \\  
    \rowcolor{highlightColor} 
    14 & 2m 6s & \underline{0.616} & \underline{0.568} & \underline{3.341} & \underline{29.19} & \textbf{4.64} & \textbf{4.86} \\ 
    20 & 2m 42s & \textbf{0.621} & 0.552 & 3.223 & \textbf{29.21} & 4.45 & 4.77 \\ 
    26 & 3m 18s & 0.603 & 0.538 & 3.165 & 29.12 & 4.33 & 4.69 \\ 
    \bottomrule
    \end{NiceTabular}
    \label{tab:views_ablation}
\end{table*}
\Cref{tab:views_ablation} reports the effect of the number of reference perspective views, $N \in \{6, 14, 20, 26\}$.
For $N=6$, we used two polar views $(\theta, \phi) = (0, \pm\frac{\pi}{2})$ and four equatorial views $(\theta, \phi) = (\frac{\pi}{2}i, 0)$ for $i \in \{0,1,2,3\}$.
For larger $N$, we used two polar views, $2L$ tilted views $(\theta, \phi) = (\frac{2\pi}{L}i, \pm\frac{\pi}{3})$, and $L$ equatorial views $(\theta, \phi) = (\frac{2\pi}{L}i, 0)$ with $L \in \{4,6,8\}$ (corresponding to $N=14,20,26$).

Using $N=6$ yields the highest scores for \textit{Imaging} and \textit{QAlign}, but substantially degrades geometric metrics such as \textit{Distortion} and \textit{Continuity}.
In this setting, reference views have little to no overlap at each diffusion step, so the optimization can produce high-fidelity local views while failing to enforce global seamlessness over the full panorama.
Increasing the number of views to $N=14$ improves \textit{Distortion} and \textit{Continuity} while maintaining strong perceptual and text alignment scores, providing the best overall trade-off between quality, consistency, and runtime in our experiments.
Further increasing $N$ beyond 14 slightly improves some metrics but tends to degrade others, while incurring higher runtime.
We conjecture that with heavier overlap, the least-squares fusion becomes more redundant and can make the iterative solve less well-conditioned, leading to diminishing returns under a fixed iteration budget.

\section{Conclusion}
We presented \method, a fast training-free framework for panoramic image synthesis that extends MultiDiffusion to arbitrary linear projections between perspective views and ERP panoramas. By casting the update under linear mappings as a regularized least-squares problem and solving it with a matrix-free Krylov method, our approach removes the direct-sampling constraint of prior training-free baselines, enabling denser and more natural mappings between panorama and perspective views. 
This reduces the number of view evaluations during denoising. 
As a result, \method improves generation quality, text alignment, and geometric consistency while achieving about 15.36$\times$ faster inference than the best-performing training-free baseline.

Beyond panoramas, this formulation suggests a general recipe for optimization-based synthesis on non-planar domains; future work includes extending it to other surface topologies (e.g., torus-like representations for walk-through content) and to texture optimization directly on mesh surfaces.

\section*{Acknowledgment}

This work was partially supported by JST Moonshot R\&D Grant Number JPMJPS2011.


%
%
\bibliographystyle{splncs04}
\bibliography{main}

\newpage

\onecolumn{ 
    \centering 
    \Large 
    \textbf{\paperTitle} \\ \vspace{0.5em}Supplementary Material \\ \vspace{1.0em} 
} 

\beginsupplement
\section{Details of the user study}
\label{sup:sec:user_study}

We conducted a user study to subjectively evaluate the visual quality and seamlessness of the generated panoramas.
We randomly sampled four prompt sets from the dataset and generated panoramas using DynamicScaler~\cite{Liu2025DynamicScaler}, SphereDiff~\cite{park2025spherediff}, and \method.
For each panorama, we also provided a 10-second video showing the scene from multiple viewing directions.

Each participant evaluated 12 panorama-video pairs in total (4 prompt sets $\times$ 3 methods) and rated every sample on the following three aspects using a 5-point Likert scale~\cite{likert1932technique} (1: Poor, 2: Subpar, 3: Fair, 4: Good, 5: Excellent):
\begin{enumerate}
    \setlength{\itemsep}{0cm}
    \item \textbf{Image Quality}: Rate the overall visual quality of the image. Lower scores should be assigned to images that are blurry, noisy, lacking detail, or exhibiting visible artifacts.
    \item \textbf{Distortion}: Rate whether the image contains geometric distortions. Lower scores should be given if objects or scene structure look unnaturally stretched or deformed. Please focus on shape and structure, not on the content or style. Pay attention to regions away from the horizontal viewing direction, such as upward or downward views.
    \item \textbf{Continuity}: Rate how smoothly and consistently the image is connected across different viewing directions. Lower scores should be assigned when visible seams, breaks, or structural misalignments are present.
\end{enumerate}

We collected 132 ratings for each aspect from 11 participants.
To assess statistical significance, we report 95\% confidence intervals (CIs) based on the standard error.
Table~2 in the main paper summarizes the user study results.
\method achieves the highest Image Quality ratings while also obtaining comparable or slightly better Distortion and Continuity ratings than SphereDiff with significantly faster inference.

\section{Details of the Krylov solvers}
\begin{algorithm}[t]
\caption{PCG solver for regularized least squares}
\label{alg:pcg_solver}
\begin{algorithmic}[1]
\Require Initial panorama feature $J^{(0)}$, stacked denoised target $d$, stacked operator $S$, adjoint operator $S^\top$, regularization operator $L$, diagonal preconditioner $M$, regularization weight $\lambda$, \# of iterations $\tau$, and tolerance $\varepsilon$

\Statex
\Statex \textbf{Phase 1: Initialization}
\State Define the linear operator $H(\cdot) \coloneqq S^\top S(\cdot) + \lambda L^\top L(\cdot)$ \Comment{Normal-equation operator}
\State $b \leftarrow S^\top d, \; J \leftarrow J^{(0)}, \; r \leftarrow b - H(J)$ 
\State $z \leftarrow M^{-1} r, \; \rho \leftarrow \langle r, z \rangle, \; p \leftarrow z$ \Comment{Preconditioner and search direction}

\Statex
\Statex \textbf{Phase 2: PCG iteration}
\For{$k = 1, 2, \dots, \tau$}
    \State $q \leftarrow H(p), \; \alpha \leftarrow \rho / \langle p, q \rangle, \; J \leftarrow J + \alpha p, \; r \leftarrow r - \alpha q$ \Comment{PCG step}
    \If{$\|r\| / \|b\| < \varepsilon$}
        \State \textbf{break} \Comment{Stop if converged}
    \EndIf
    \State $z \leftarrow M^{-1} r, \; \beta \leftarrow \langle r, z \rangle / \rho, \; \rho \leftarrow \langle r, z \rangle, \; p \leftarrow z + \beta p$ \Comment{Update search direction}
\EndFor

\State \Return $J$ \Comment{Return the optimized panorama feature}
\end{algorithmic}
\end{algorithm}

\begin{algorithm}[t]
\caption{LSMR solver in \method}
\label{alg:lsmr_solver}
\begin{algorithmic}[1]
\Require Initial panorama feature $J^{(0)}$, augmented stacked denoised target $\tilde{d}$, augmented operator $\tilde{S}$, adjoint operator $\tilde{S}^\top$, \# of iterations $T$, and tolerance $\varepsilon$

\Statex
\Statex \textbf{Phase 1: Initialization}
\State $J \leftarrow J^{(0)}, \; r \leftarrow \tilde{d} - \tilde{S} J$
\State $\beta \leftarrow \|r\|, \; u \leftarrow r / \beta, \; \alpha \leftarrow \|\tilde{S}^\top u\|, \; v \leftarrow \tilde{S}^\top u / \alpha$ \Comment{Bidiagonalization}
\State $\mathcal{S} \leftarrow \mathrm{InitializeLSMRState}(\alpha, \beta)$ \Comment{LSMR auxiliary state}

\Statex
\Statex \textbf{Phase 2: LSMR iteration}
\For{$k = 1, 2, \dots, T$}
    \State $\beta \leftarrow \|\tilde{S} v - \alpha u\|, \; u \leftarrow (\tilde{S} v - \alpha u) / \beta$
    \State $\alpha \leftarrow \|\tilde{S}^\top u - \beta v\|, \; v \leftarrow (\tilde{S}^\top u - \beta v) / \alpha$
    \State $(\Delta J, \mathcal{S}) \leftarrow \mathrm{LSMRUpdate}(\mathcal{S};\ \alpha, \beta, v)$ \Comment{compute direction w/ state update}
    \State $J \leftarrow J + \Delta J,\ r \leftarrow \tilde{d} - \tilde{S} J$ \Comment{LSMR step}
    \If{$\|r\| / \|\tilde{d}\| < \varepsilon$}
        \State \textbf{break} \Comment{Stop if converged}
    \EndIf
\EndFor

\State \Return $J$ \Comment{Return the optimized panorama feature}
\end{algorithmic}
\end{algorithm}

We evaluated both PCG~\cite{Hestenes1952CG} and LSMR~\cite{Fong2011LSMR} for solving Eq.~(8).
\Cref{alg:pcg_solver,alg:lsmr_solver} summarize the update rules of the PCG and LSMR solvers, respectively.
For PCG, we solve the regularized normal equation using the linear operator
$H(\cdot) = S^\top S(\cdot) + \lambda L^\top L(\cdot)$
together with a diagonal preconditioner.
The diagonal preconditioner is derived from the squared interpolation weights corresponding to $S^\top S$.
For LSMR, we instead solve the equivalent augmented least-squares problem.
Here, $\mathcal{S}$ denotes the auxiliary recurrence state maintained by LSMR, including the rotation coefficients and search-direction-related variables.
In both solvers, $S$, $S^\top$, $\tilde{S}$, and $\tilde{S}^\top$ are implemented as matrix-free operators, avoiding explicit matrix instantiation in memory.
\section{Details of the VLM-based evaluation}

To evaluate panoramic seamlessness, specifically Distortion and Continuity, we used Qwen2.5-VL~\cite{bai2025qwen25vltechnicalreport}, an open-source vision-language model.
\Cref{tab:vlm_prompt} shows the evaluation prompt.
Our evaluation protocol follows the VLM-based procedure introduced in SphereDiff~\cite{park2025spherediff}.
However, we found that, when using the original prompt, the model tended to assign lower scores based on image content or artistic style rather than purely geometric criteria.
We therefore explicitly instructed the model to ignore content and style, focusing only on geometric consistency.

\begin{table*}[!t]
\centering
\renewcommand{\arraystretch}{1.2}
\caption{\textbf{Evaluation Prompt for Qwen-VL}. Evaluation prompt used with Qwen2.5-VL~\cite{bai2025qwen25vltechnicalreport} for assessing distortion and continuity in generated panoramas. The prompt follows SphereDiff~\cite{park2025spherediff} with slight modifications to better isolate geometric consistency from image content and style.}
\label{tab:vlm_prompt}
\begin{tabular}{|p{0.95\linewidth}|}
\hline
You are an evaluator assessing an image generation model on a single-image basis. Your evaluation is based on the following two criteria: \\
\\
1. Distortion: Evaluate only geometric distortion (not content/style). High if it could pass as a normal camera photo; low if you notice stretching/warping/bending or inconsistent proportions. \\
2. Continuity: Evaluate only continuity (not content/style). Score low if there is any visible break anywhere, including cut, misalignment, tearing, duplicated edges, or abrupt texture/lighting change. \\
\\
Each criterion is rated on a five-point scale: Excellent (5), Good (4), Fair (3), Subpar (2), and Poor (1). You will receive one image at a time. For each criterion, provide a concise reason for the score before listing the rating. \\
\\
Format your response as follows: \\
- Distortion: (Brief reason) → Score \\
- Continuity: (Brief reason) → Score \\

\hline
\end{tabular}
\end{table*}

\section{Experiments with SANA}
\begin{table*}[t]
\small
    \centering
    \caption{Panorama generation results with SANA~\cite{xie2025sana}.}
    \begin{NiceTabular}{l@{\hspace{6pt}}c@{\hspace{6pt}}c@{\hspace{6pt}}c@{\hspace{6pt}}c@{\hspace{6pt}}c@{\hspace{6pt}}c@{\hspace{4pt}}c}
    \toprule
    Method & Runtime & Aesthetic & Imaging & QAlign & CLIP & Distortion & Continuity \\
    \midrule
    SphereDiff & 4m 11s & \textbf{0.550} & 0.531 & 2.985 & \textbf{27.42} & \textbf{3.62} & 4.15 \\ 
    \rowcolor{highlightColor} Ours & \textbf{34s} & 0.538 & \textbf{0.566} & \textbf{3.001} & 26.67 & 3.38 & \textbf{4.26} \\
    \bottomrule
    \end{NiceTabular}
    \label{tab:sana_baseline_comparison}
\end{table*}

In addition to FLUX~\cite{flux2024}, we evaluated \method with SANA~\cite{xie2025sana} and compared it against SphereDiff.
\Cref{tab:sana_baseline_comparison} presents the results for SANA used as the base T2I model.
Because SANA is a computationally efficient text-to-image model with a highly compressed latent space and a linear diffusion transformer, its inference is reported to be approximately 100$\times$ faster than FLUX.
As a result, the cost of each MultiDiffusion step is drastically reduced, increasing the relative runtime contribution of the linear solver.
Even in this setting, \method achieves a substantial speedup ($7.38\times$) while maintaining comparable performance.

\begin{table}[t]
    \small
    \centering
    \caption{Quantitative comparison on text-to-panoramic video generation. Motion denotes motion smoothness in VBench.}
    \begin{NiceTabular}{l@{\hspace{6pt}}c@{\hspace{6pt}}c@{\hspace{6pt}}c@{\hspace{6pt}}c@{\hspace{6pt}}c@{\hspace{6pt}}c}
    \toprule
    {\footnotesize Method} & {\footnotesize Runtime$\downarrow$} & {\footnotesize QAlign$\uparrow$} & {\footnotesize CLIP$\uparrow$} &{\footnotesize Distortion$\uparrow$} & {\footnotesize Continuity$\uparrow$} & {\footnotesize Motion$\uparrow$} \\
    \midrule
    \multicolumn{7}{l}{\rowcolor{headerColor} {\textbf{Training-based panoramic video generation}} \hfill} \\
    360DVD & \deemph{1m 31s} & \underline{2.36} & 24.13 & \underline{2.98} & \textbf{3.41} & 0.32  \\
    ViewPoint & \deemph{2m 54s} & 1.61 & 24.39 & 1.89 & 2.07 & \textbf{0.66}  \\
    \midrule
    \multicolumn{7}{l}{\rowcolor{headerColor} {\textbf{Training-free panoramic video generation}} \hfill} \\ 
    SphereDiff & 25m 14s & \textbf{2.67} & \underline{27.72} & 2.70 & 3.13 & 0.43 \\ 
    \rowcolor{highlightColor} Ours & \textbf{2m 44s} & \underline{2.36} & \textbf{28.19} & \textbf{3.01} & \underline{3.37} & \underline{0.44} \\
    \bottomrule
    \end{NiceTabular}
    \label{sup:tab:video_results}
\end{table}

\section{Panoramic Video Generation}

In addition to static panorama generation, we adapted our method to text-to-panoramic video generation.
We used LTX-Video~\cite{HaCohen2024LTXVideo} as a pre-trained text-to-video generator.
For the baselines, we compared against 360DVD~\cite{Wang2024360DVD} and ViewPoint~\cite{Fang2025ViewPoint} as training-based baselines, and SphereDiff, which uses the same video model, as a training-free baseline.
Note that DynamicScaler was omitted because its official implementation only supports image-to-video generation (from static panorama to dynamic panorama) and does not support text-to-video generation.
Following each default setup, the output sizes are $16 \times 512 \times 1024$ for 360DVD, $ 49 \times 512 \times 1024$ for ViewPoint, and $121 \times 1024 \times 2048$ for SphereDiff and ours. 
Thus, the training-based runtimes are only references due to their different output sizes, whereas the SphereDiff vs. ours runtime comparison is direct.
\Cref{sup:tab:video_results} shows that our method reduces the runtime over SphereDiff from 25m 14s to 2m 44s, while achieving better scores on most metrics except comparable QAlign score. 
These results demonstrate that our training-free formulation transfers to panoramic video generation with a clear efficiency gain.
\section{Latent and projection variants}

In this section, we present results with alternative latent representations and projection variants to demonstrate the generality of \method.
In addition to the ERP latent and the bilinear projection between panorama and perspective views used in the main paper, we also evaluate a spherical latent representation and a splatting-based forward projection from 3D points to perspective views.
For clarity, in this section, we refer to the bilinear projection used in the main paper as a backward projection (Bwd) when comparing it with the new forward-projection variant (Fwd), since the key difference lies in the projection direction.

\subsection{Spherical latent representation}

Inspired by SphereDiff~\cite{park2025spherediff}, we represent a 360-degree panoramic scene using points distributed on the surface of a unit sphere.
Specifically, we define $J$ as a set of $N_s$ points on the unit sphere.
Each point has a 3D position $(x_i, y_i, z_i)$ and a $c$-dimensional feature vector, where $c$ matches the latent dimension of the base model.
We initialize the point positions using the Fibonacci lattice~\cite{Hardin2016fibonaccisphere} to distribute them approximately uniformly over the sphere surface.
We denote this spherical latent representation as SPH.

With SPH, we tested both the backward projection introduced in Eq.~(11) of the main paper and the forward projection described in the following section.
For the backward projection, we used the four neighboring points as $\mathcal{N}(i_\mathbf{u}, j_\mathbf{u})$, determined by the distances between the back-projected 3D point on the sphere corresponding to each perspective pixel $(i_\mathbf{u}, j_\mathbf{u})$ and the $N_s$ sphere points.

\subsection{Splatting-based forward projection from 3D points to perspective views}

As a variant of the projection from a 360-degree scene to perspective views, we implement a forward projection (Fwd) by splatting all 3D points onto each perspective view and computing interpolation weights based on distances on the perspective image plane.
Specifically, each 3D point $\mathbf{x}_p$ is first projected onto a perspective view using Eq.~(9), yielding a continuous pixel location $(u_{p, M_{t,i}}, v_{p, M_{t,i}})$.
We then scatter its feature to the neighboring pixels
\[
\mathcal{N}_p(u_{p, M_{t,i}}, v_{p, M_{t,i}})
=
\{(\floor{u_{p, M_{t,i}}} + \delta_u, \floor{v_{p, M_{t,i}}} + \delta_v) \mid \delta_u, \delta_v \in \{0, 1\}\}
\]
with weights defined by $\kappa(u - u_{p, M_{t,i}})\kappa(v - v_{p, M_{t,i}})$.
This projection differs from the backward projection used in Sec.~3.3 in that every 3D point can contribute to perspective pixels, provided that the set of perspective views covers all directions.

In addition to the spherical latent representation, we evaluated this splatting-based forward projection with the ERP latent described in the main paper.

\subsection{Performance comparison of latent and projection variants}
\begin{table*}[t]
\small
    \centering
    \caption{Panorama generation results with different latent and projection designs.}
    \begin{NiceTabular}{c@{\hspace{2pt}}c@{\hspace{2pt}}c@{\hspace{6pt}}c@{\hspace{6pt}}c@{\hspace{6pt}}c@{\hspace{6pt}}c@{\hspace{6pt}}c@{\hspace{4pt}}c}
    \toprule
    Latent & Projection & Runtime & Aesthetic & Imaging & QAlign & CLIP & Distortion & Continuity \\
    \midrule
    ERP & Fwd & 4m 42s & 0.609 & 0.555 & 3.22 & 28.72 & 4.47 & 4.77 \\
    \rowcolor{highlightColor} 
    ERP & Bwd & \textbf{2m 6s} & 0.616 & \textbf{0.568} & \textbf{3.34} & \textbf{29.19} & \textbf{4.64} & \textbf{4.86} \\
    SPH & Fwd & 5m 38s & 0.605 & 0.552 & 3.24 & 28.78 & 4.36 & 4.71 \\
    SPH & Bwd & 2m 20s & \textbf{0.631} & \textbf{0.568} & 3.27 & 29.03 & 4.57 & 4.85 \\
    \bottomrule
    \end{NiceTabular}
    \label{sup:tab:latent_and_proj_variants}
\end{table*}

\Cref{sup:tab:latent_and_proj_variants} compares the performance of different latent and projection variants.
For the spherical latent representation, we used $N_s = 100{,}000$.
Because forward projection requires projecting all $N_s$ points into each perspective view at every update, it is slower than backward projection.
It also slightly degrades all evaluation scores.
We conjecture that forward projection induces denser overlap among neighboring pixels than backward projection, potentially causing excessive averaging that leads to overly smooth or blurred representations.

For the spherical latent representation, we observed improvement in the Aesthetic score while maintaining comparable performance on the other metrics.
Considering both runtime and overall performance, we adopt ERP with backward projection as the default setting.

More importantly, these results show that \method generalizes across different latent representations and projection operators.
In particular, it can be applied not only to the ERP latent used in DynamicScaler-like settings but also to the spherical latent representation introduced in SphereDiff, providing a unified framework across these design choices.
Also, our additional experiments suggest that using a larger $N_s$ in SPH and a higher panorama resolution in ERP can further improve the overall scores.
These results suggest that further gains may be obtained by making better choices of the latent representation and the projection operator, which we leave for future work.
\section{Sensitivity study on the post-refinement timesteps}
\Cref{sup:tab:stopping_time_step_ablation} shows a sensitivity study on the stopping timestep $T_M$.
Panorama generation without \method ($T_M=0$) produces disconnected panoramas, while image quality and text alignment are degraded without view-wise post refinement ($T_M=28$).
Intermediate values perform better overall, and we choose $T_M=23$ for its balanced performance.
This supports the importance of both the \method and the post-refinement stage.

\begin{table}[t]
    \small
    \centering
    \caption{Sensitivity study for the stopping time step $T_M$.}
    \begin{NiceTabular}{l@{\hspace{6pt}}c@{\hspace{6pt}}c@{\hspace{6pt}}c@{\hspace{6pt}}c}
    \toprule
    $T_M$ & QAlign $\uparrow$ & CLIP $\uparrow$ & Distortion $\uparrow$ & Continuity $\uparrow$ \\
    \midrule
    0 (No \method) & 2.23 & 18.79 & 1.60 & 1.68 \\
    5 & 3.09 & 27.51 & 3.82 & 4.43 \\
    10 & 3.35 & 28.96 & 4.48 & 4.76 \\
    15 & \textbf{3.38} & \textbf{29.25} & 4.52 & 4.83 \\
    20 & 3.34 & 28.74 & 4.60 & 4.85 \\
    \rowcolor{highlightColor} 23 & 3.34 & 29.19 & \textbf{4.64} & \textbf{4.89} \\
    25 & 3.34 & 29.06 & 4.62 & 4.80 \\
    28 (No view-wise post refinement) & 2.74 & 26.80 & 4.39 & 4.60 \\
    \bottomrule
    \end{NiceTabular}
    \label{sup:tab:stopping_time_step_ablation}
\end{table}
\section{Additional generated samples}
\begin{figure}[t]
    \centering
    \includegraphics[width=\linewidth]{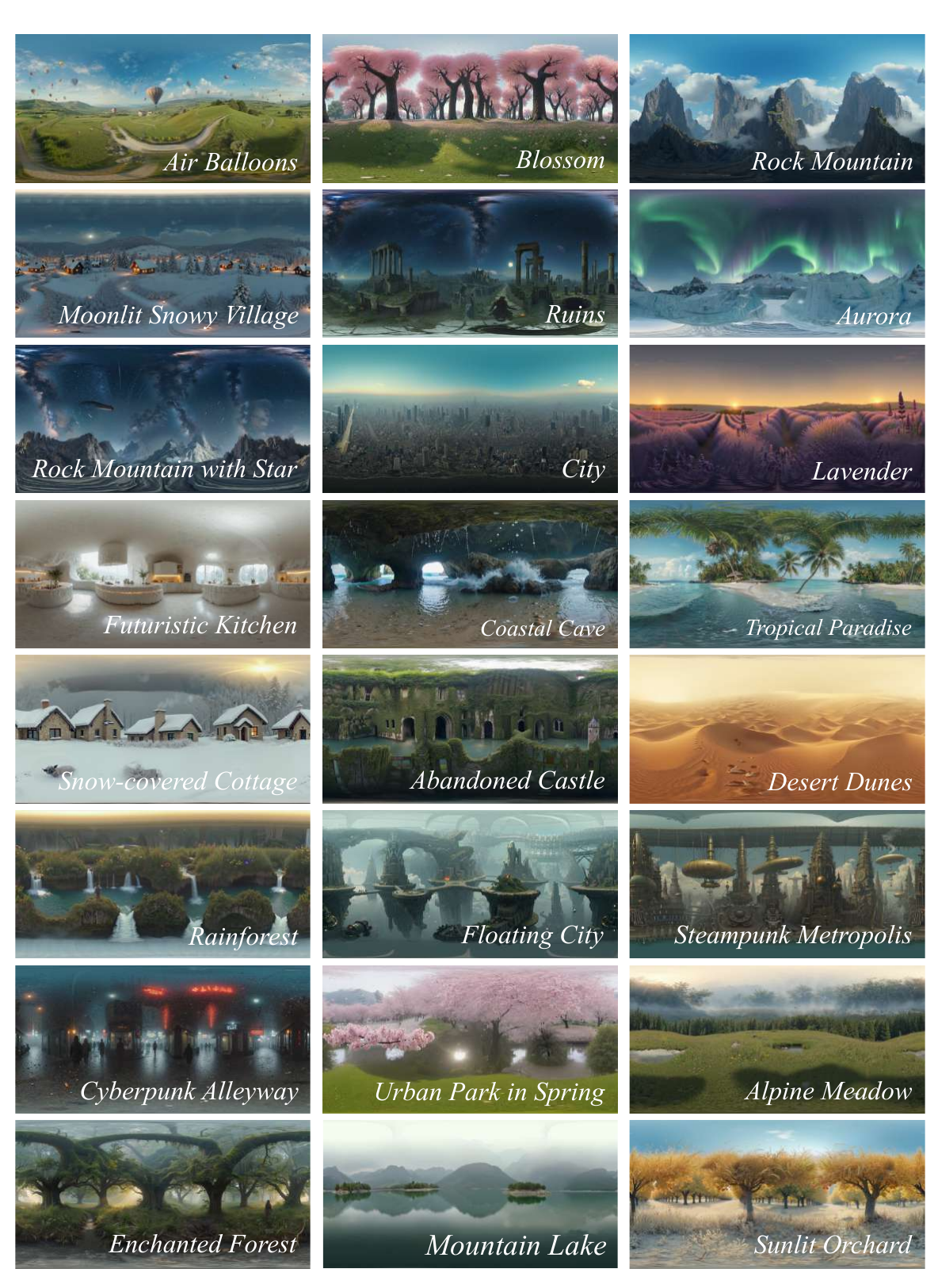}
    \caption{Generated panoramas by \method.}
    \label{sup:fig:pano_gallery}
\end{figure}

\Cref{sup:fig:pano_gallery} presents additional panorama samples generated by \method.
For the text prompts, we followed SphereDiff~\cite{park2025spherediff} and prepared three prompts per scene, each describing the upper, horizontal, and lower parts of the scene.
These text prompts are automatically generated using Qwen3.5-2B~\cite{qwen3.5} based on the scene labels shown in the figure.
We further provide a demo in the supplementary material for comparison with other baselines, including both panoramas and their corresponding perspective-view videos.
A more detailed comparison is available on our project page: \url {https://ahykw.github.io/lfmd/}.

\end{document}